# Universal Language Modelling agent

## ULMa

Anees Aslam [1]

Ulma.model@gmail.com | aneestags786@gmail.com

https://www.linkedin.com/in/anees-aslam/

## Abstract

*Large Language Models are designed to understand complex Human Language. Yet, Complexity of animal language has long intrigued researchers striving to bridge the communication gap between humans and other species. This research paper introduces a novel approach that draws inspiration from the linguistic concepts found in the Quran, a revealed Holy Arabic scripture dating back 1400 years. By exploring the linguistic structure of the Quran, specifically the components of ism, fil, and harf, we aim to unlock the underlying intentions and meanings embedded within animal conversations using audio data.*

*To unravel the intricate complexities of animal language, we employ word embedding techniques to analyse each distinct frequency component. This methodology enables the identification of potential correlations and the extraction of meaningful insights from the data. Furthermore, we leverage a bioacoustics model to generate audio, which serves as a valuable resource for training natural language processing (NLP) techniques. This Paper aims to find the intention\* behind animal language rather than having each word translation.*

## 1    Introduction

Understanding animal language has been a longstanding pursuit for researchers seeking to unravel the mysteries of animal communication and bridge the gap between humans and other species. While progress has been made in deciphering certain aspects of animal vocalizations, a comprehensive understanding of their intentions and the intricacies of their language remains elusive. This paper proposes a novel approach that draws inspiration from the linguistic patterns found in the Quran, a holy Arabic scripture revealed over 1400 years ago. By exploring the linguistic concepts presented in the Quran and their potential resonance with animal conversation, we aim to shed light on the underlying structure and intentions of animal communication.

Traditional studies on animal communication have largely focused on deciphering vocalizations, gestures, and other forms of non-verbal communication. However, these approaches often fall short in capturing the richness and complexity of animal language. Inspired by the Quran's linguistic framework, we introduce the concept of mapping animal communication patterns into three linguistic components: ism, fil, and harf. These components draw parallels to the Quranic linguistic categories and provide a structured framework for analysing and interpreting animal communication.

---

*The Term " **Intention**" in this paper refers to the underlying meaning or purpose behind their vocalizations. It focuses on understanding the target object or concept being conveyed by the animals and their likelihood or inclination towards it.*

To unravel the intentions encoded within animal communication, we leverage techniques commonly used in natural language processing. Word embedding, a powerful method in NLP, is applied to each of the independent frequency components, enabling the extraction of semantic relationships and underlying patterns. By utilizing word embedding techniques, we aim to decipher the subtle nuances and intentions encoded within the animal language.

In addition to linguistic analysis, we incorporate a bioacoustics model to generate audio data that reflects the patterns and characteristics of animal communication. This audio data serves as valuable input for training an advanced natural language processing technique. Employing deep learning algorithms, the NLP technique is fine-tuned using the generated animal audio and enriched with human feedback, fostering an iterative learning process that refines the understanding and interpretation of animal intentions.

However, this approach is not without its challenges. Validating the mapping of animal communication patterns to linguistic components inspired by the Quran requires extensive analysis and validation. The availability and quality of animal audio data, which are crucial for training the NLP technique, pose significant hurdles in terms of data collection and labelling. Furthermore, ethical considerations surrounding the treatment and welfare of animals during the research process demand careful attention and adherence to responsible research practices.

This paper, presents an initial exploration of our proposed approach, emphasizing its potential to enhance our understanding of animal language. While the use of large language models has proven successful in understanding human language, the application of these models to decipher animal communication remains relatively unexplored. By combining insights from linguistics, bioacoustics, and natural language processing techniques, our approach holds promise in unravelling the intentions and intricacies of animal language. Through further research and validation, the aim is to unlock the mysteries of animal communication and forge a deeper connection with the non-human species that share our planet.

*"While Traditional Large Language approach are designed in understanding complex human languages as a result of thousand years of civilization. Understanding Animal Intentions takes us back in time to get into the bare bone elements of language. The Quran miraculous way of communicating rich content with understandable pattern acts as bridge. "*

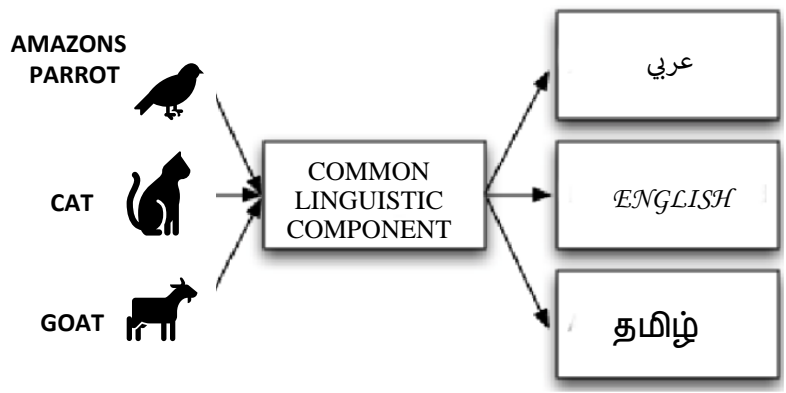

Figure 1.1: The basic interpretation of ULM model

## 2       Pattern Identification

### 2.1   Linguistics references

In the context of our NLP approach, [1] we utilize three linguistic building blocks: Ism, Fil, and Harf. To draw a comparison with the English language literature, we can explain these components as follows:

Ism: Ism is a superset of Noun, encompassing names, objects, and other elements that carry standalone meaning. In other words, Ism represents words that function as nouns in a sentence.

Fil: Fil can be understood as a counterpart to the concept of a verb. It represents action words and encompasses verbs with an additional correlation to the factor of time.

Harf: Harf refers to words that do not possess individual meanings but serve the purpose of providing context to the sentence and the words within it. Harf aids in understanding the relationships and connections between different elements in a sentence

### 2.2   Parallelization with Animal Audio

Continuing the discussion from the previous point, we can draw parallels between these linguistic building blocks and the study of animal intention. By examining the fundamental elements of language and their role in animal communication, we may gain insights into the intentions of animals. While animals do not possess a language system comparable to human languages, they do communicate through various signals, vocalizations, and behaviours. By identifying analogous components in animal communication, we can better understand their intentions and motives.

It's important to note that this parallelization may require scientific research and observation across different animal species. While the linguistic building blocks provide a framework for understanding human language, their application to animal communication should be approached with care and supported by empirical evidence.

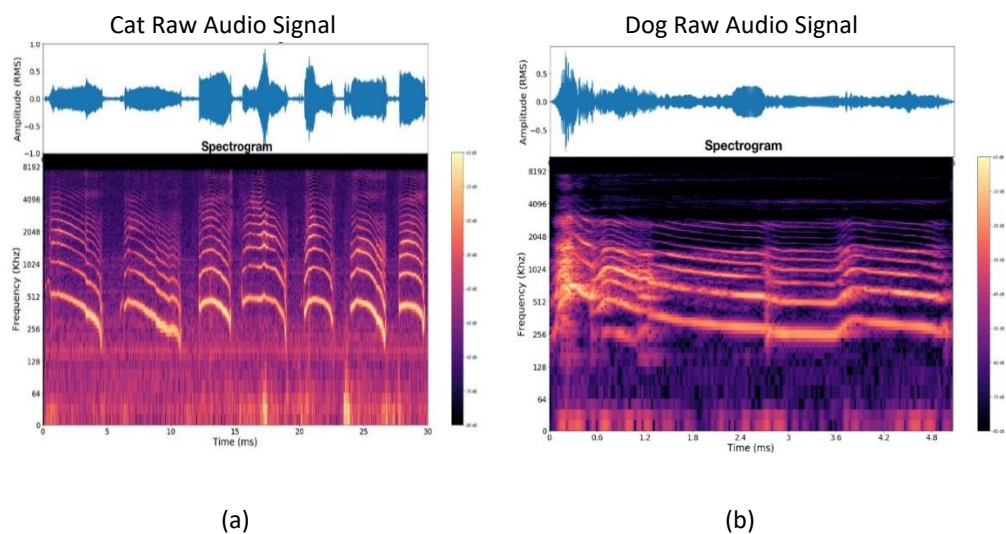

(a)                  (b)

Figure 2.1: Spectrogram representation of (a) Cat Meow and (b) Dog Bark

## 2.3    Identification and Breakdown of Vocal Pattern

In analysing the vocal patterns of different animals, it is observed that they possess a pattern consisting of three main components: Ism, Fil, and Harf.

*This breakdown can be visualized as follow*

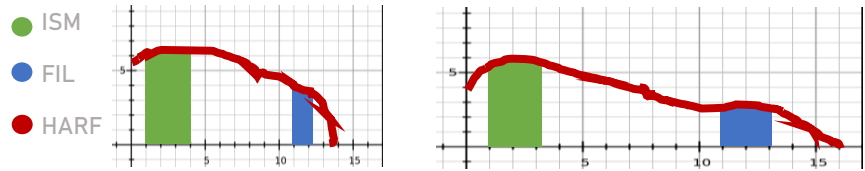

Figure 2.2:  Depicting the identification of audio patterns for cats and dogs

The primary focus of the vocal pattern is the Ism, which represents the main intention or message conveyed by the animal. Following the Ism, there is a secondary component known as Fil, which provides additional information and context to the vocalization. The Harf component can be seen as a string that is mapped over the entire timeline, contributing to the overall structure and flow of the vocal pattern.

The distance between the Ism and Fil components serves as a characteristic that distinguishes different animals. This distance may vary, highlighting unique aspects of each species' vocal communication. The variation in height and density of the Ism and Fil components within the vocal pattern represents the intention or motive of the target animal. By analysing these variations, we can gain insights into the animal's purpose or state.

It's important to note that the presence of Fil in the vocalization is not compulsory. Some vocal patterns may only consist of the Ism component. However, the Ism is essential for conveying meaning, and without it, the vocalization would not hold any significance. The Harf component, although not individually meaningful, contributes to the overall structure and context of the vocal pattern.

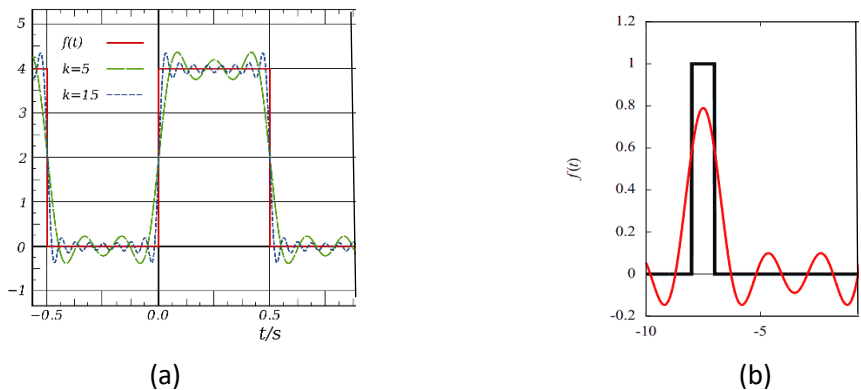

Figure 2.3: (a) Signal extraction of ISM (b) Signal extraction of FIL

# 3    Nonlinear Language Construction using Ism, Fil, and Harf

Based on our previous discussion on the construction of animal language components (Ism, Fil, and Harf), we can propose a scientific theory that highlights the role of these components in creating nonlinearity and facilitating a coherent flow in animal vocalizations.

*a  Nonlinearity as a Language Feature*

The theory posits that animal vocalizations exhibit nonlinearity, which refers to the deviation from a linear or sequential structure. Nonlinearity in animal communication arises from the dynamic interplay between Ism, Fil, and Harf components.

*b  Ism and Fil Placement*

The Ism component represents the primary intention or message conveyed by the animal, while the Fil component captures the relationship between the Ism and the specific animal's perception or response. In constructing animal vocalizations, the placement of Ism and Fil within the sequence or timeline introduces nonlinearity. This nonlinearity can manifest in the form of temporal variations, pauses, or changes in the order or emphasis of Ism and Fil components.

*c  Harf as a Connector*

The Harf component, which acts as a contextual word or connector, plays a crucial role in creating a smooth and coherent flow within animal vocalizations. Harf serves as the "string" that ties the two ends of the timeline together, facilitating transitions and linking Ism and Fil components. By using Harf strategically, animals can create nonlinearity while maintaining a cohesive and understandable communication pattern.

*d  Normalization and Flow*

To ensure effective communication, normalization techniques are employed. Normalization involves establishing patterns and consistent associations between Ism, Fil, and Harf components. This process enhances the flow and coherence of animal vocalizations, enabling better comprehension and interpretation.

This scientific theory suggests that animal vocalizations are constructed through the nonlinear arrangement of Ism and Fil components, connected by Harf. By understanding and analysing these components, researchers can unravel the intricate language systems of different animal species and gain insights into their intentions, relationships, and overall communication patterns. It's important to note that this theory is a hypothesis proposed based on our discussion and may require further empirical research and data analysis to validate its claims. The complexity of animal communication and the diversity of species necessitate ongoing scientific exploration to refine and expand our understanding of this intriguing field.

$$ULM = Harf * f(Ism + Chirps + Fil)$$

"*Chirps*" represents units (distance) between Ism and Fil.

## 3.1 Construction of ISM

The Ism, representing the key intention in animal vocalizations, is constructed based on results obtained from supervised learning. To accomplish this, we engage in interactive sessions with a specific group of animals, capturing their audio signals. These signals are comprised of multiple frequencies, which we then map to create greyscale images for each data sample. We apply word embedding techniques to encode the information contained within these images.

To extract Ism wave from a noise signal, you can use a process called signal conditioning or filtering. One common method is to use a comparator Design [4]. Equation that describes the process of extracting a Ism wave from a noisy signal using a comparator:

*Vout(t) = (Vmax - Vmin) * sign(Vin(t) - Vth) + (Vmax + Vmin) / 2*

- Vout(t) is the output square wave signal at time t.
- Vin(t) is the input noise signal at time t.
- Vth is the threshold voltage of the comparator, which determines when the signal transitions from high to low or vice versa.
- Vmax and Vmin are the desired maximum and minimum voltages of the output square wave, respectively.

The "sign" function returns -1 for negative inputs, 0 for zero inputs, and 1 for positive inputs. This function ensures that the output signal transitions between Vmin and Vmax based on the input signal's voltage compared to the threshold voltage. Subsequently, we test and train the model using labelled data.

By collecting numerous examples over several months, we aim to strengthen the probability percentage associated with the relationship between the Ism and the specific intention it represents in the cat's vocalization.

The positional encodings have the same dimension $d_{model}$ as the embeddings, so that the two can be summed. There are many choices of positional encodings, learned and fixed. In this work, we use sine and cosine functions of different frequencies:

$$PE_{(pos,2i)} = sin(pos/10000^{2i/d_{model}})$$

$$PE_{(pos,2i+1)} = cos(pos/10000^{2i/d_{model}})$$

where $pos$ is the position and $i$ is the dimension. That is, each dimension of the positional encoding corresponds to a sinusoid. The wavelengths form a geometric progression from $2\pi$ to $10000\cdot2\pi$. We chose this function because we hypothesized it would allow the model to easily learn to attend by relative positions. The above points outline a specific approach for constructing the Ism component based on supervised learning with a particular group of animals.

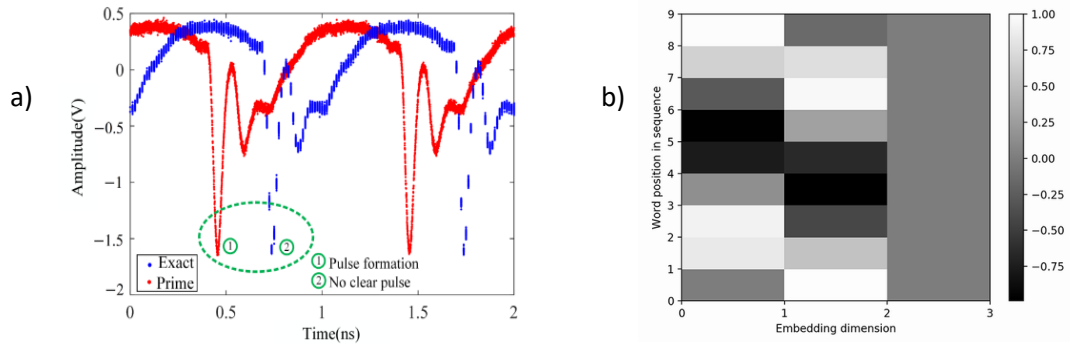

Figure 3.1: a) Extraction of Ism b) Embedding of words in dimensional vector space

## 3.2  Construction of FIL

When constructing the Fil component in the universal language model, it is essential to consider how it relates to the Ism and how the animal perceives and interacts with that Ism.

*Relationship between Ism and Fil*

The Fil component represents the relationship between the Ism and the specific animal. It captures how the animal interprets and responds to the Ism in its vocalization. The Fil component is dynamic and can change based on the animal's experiences and interactions.

In constructing the Fil component, it is crucial to take into account the individual experiences and interactions of each animal. By repeatedly observing and analysing the animal's vocalizations and behaviours, researchers can identify patterns and changes in the Fil component, allowing for a deeper understanding of the animal's intentions and their relationship to the Ism.

Relationship between the height of the Fil component (relative to Ism) and the contextual reactions or responses exhibited by animals in their vocalizations. Specifically, we aim to determine if variations in Fil height can provide insights into positive or negative reactions towards the Ism, as well as the presence or absence of interest in the subject.

Identification of patterns and correlations between Fil height ratios and contextual reactions towards Ism. A potential framework for understanding and interpreting animal vocalizations based on the Ism-Fil relationship and contextual cues.

Determination of threshold values or ranges that indicate positive or negative reactions, as well as the absence of interest.  Q = f(Ism) K = f(fil) T=f(chirps) V=f(harf)

$$\text{Threshold } (Q, K, V) = \text{softmax} \left( \frac{QK^T}{\sqrt{d_k}} \right) V$$

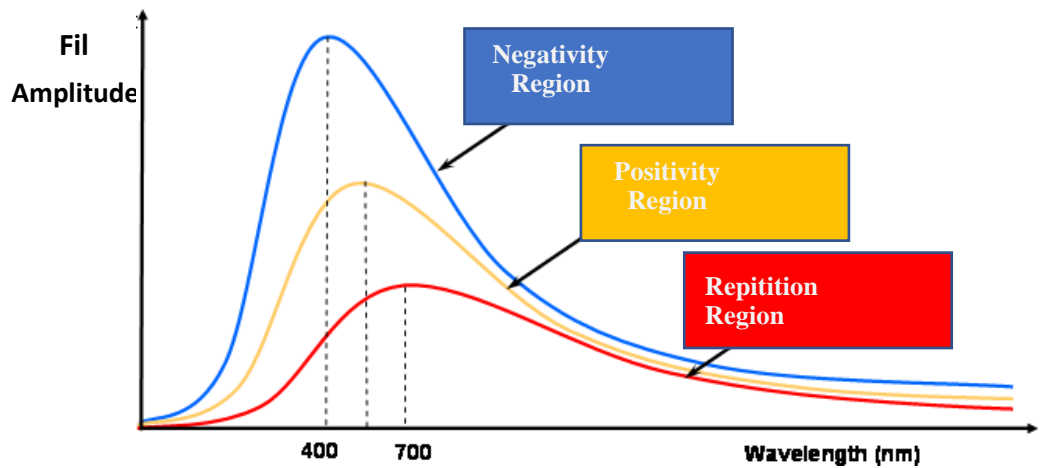

Figure 3.2: Representation of Fil Amplitude map over intention

### 3.3 Construction of HARF

Harf component in animal language can be metaphorically understood as a string that is attached to two ends of a timeline, encompassing the Ism and Fil components. Its primary function is to provide continuity, smoothness, and coherence to the communication flow.

If we consider "Ism" and "Fil" as two edges of a bridge and "Harf" as the string that connects them, we can draw an analogy to the suspension bridge system. In this analogy, "Ism" and "Fil" represent the supporting cables of the bridge, and "Harf" acts as the suspension cable that holds them together, [5] the parabolic curve of the main harf in the middle span calculated by the

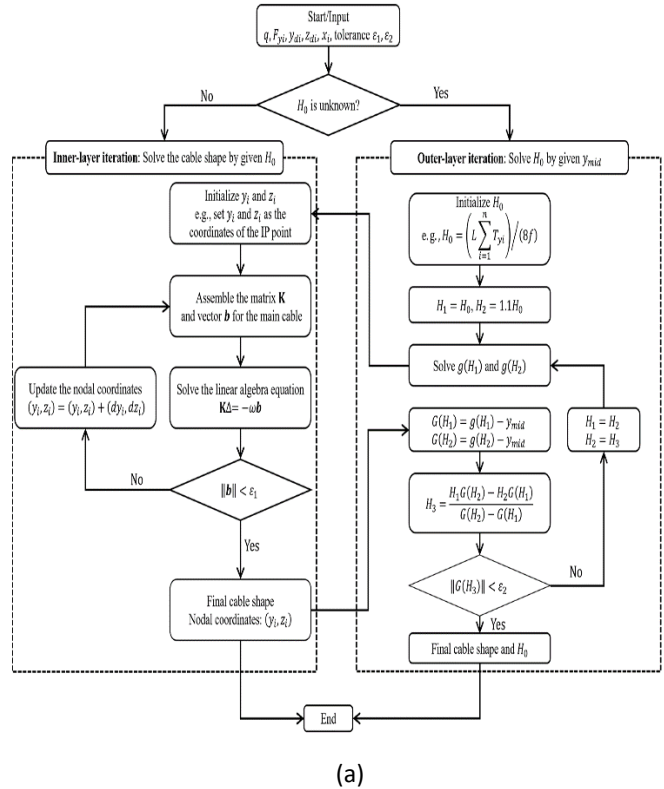

(a)

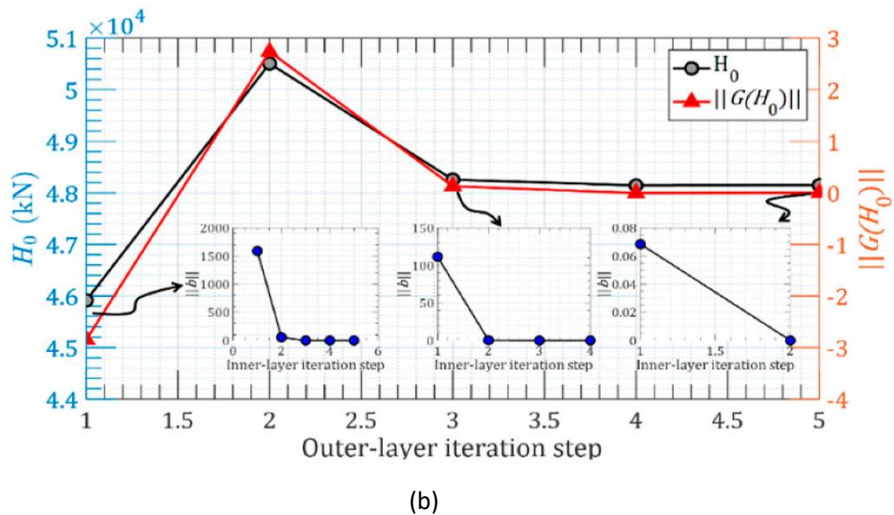

(b)

Figure 3.3: a) Flowchart of the iteration scheme for construction of harf b) Computer generation of harf component with ism and fil as inputs

### 3.4 CATGPT

To overcome insufficiency of data, I developed an natural language processing app based on Transformers [2] called CATGPT, the main objective is to interpret and respond to the actions and behaviours of cats. Through an intuitive interface, pet owners can input their cat's actions or vocalizations, and CATGPT will generate appropriate responses, simulating a conversation between the owner and their cat. It was trained on cats Dataset from **Kaggle**.

# 4 Methodology

This research builds upon the HuBERT[model [6], a self-supervised audio representation model primarily developed for human speech. We adopt a two-step methodology: HuBERT pretraining and transfer learning to bioacoustics.

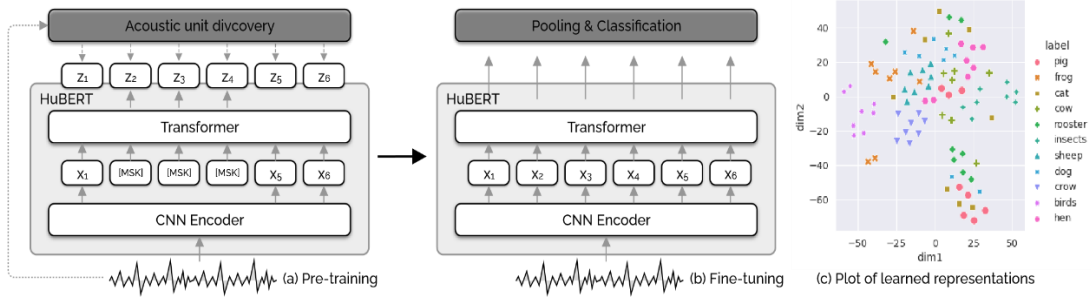

Figure 4.1 (a) pretraining and (b) fine-tuning, and (c) t-SNE plot of learned representations

*i) Data Collection*

Gather a diverse dataset of animal vocalizations encompassing various species, contexts, and Ism-Fil combinations. Ensure that the dataset includes vocalizations where the Ism and Fil components are distinguishable and measurable.

*ii) Height Measurement and Analysis*

Develop a method for accurately measuring the height of the Fil component relative to the Ism in the vocalizations. Apply mathematical techniques, such as waveform analysis or frequency-based measurements, to quantify the heights of Ism and Fil.

*iii) Contextual Annotation and Reactions*

Annotate the vocalizations with contextual information, such as the subject matter, known behavioural responses, or observed environmental cues Identify positive, negative, or neutral contextual reactions associated with the Ism-Fil combinations.

*iv) Analysis and Correlation:* Analyze the relationship between Fil height and the contextual reactions towards Ism.

## 4.1 HuBERT Pretraining

HuBERT addresses the challenge of working with continuous, raw audio waveforms lacking explicit discrete units such as phonemes or words. To overcome this, HuBERT employs acoustic unit discovery, which involves identifying discrete sound representation tokens from the raw audio data. In the initial stage of HuBERT training, cluster labels are obtained through k-means clustering applied to 39-dimensional Mel-frequency cepstral coefficient (MFCC) features.

The HuBERT architecture comprises a CNN encoder and a transformer encoder. The CNN encoder processes the raw audio waveform, generating continuous audio representations at a rate of 50 frames per second. The transformer encoder takes a masked sequence of audio representations, producing hidden representations. These hidden representations are used for predicting the acoustic units. HuBERT is pretrained using a BERT-like masked language objective, training the model to predict the acoustic units at masked positions. The model's parameters are optimized using a softmax function, comparing the similarity between the projected model output and an embedding vector for each class.

To improve the quality of discovered acoustic units, we repeat the clustering process on features extracted from an internal layer (6th for base, 12th for large architectures) of the pretrained model. The k-means algorithm is applied during this second training stage. Although it's possible to repeat this process more than twice, for our research, we employ the model from the second iteration.

## 4.2    Transfer Learning to Bioacoustics

In this phase, we leverage the learned representations from the HuBERT model for two tasks: classification and detection. Classification involves assigning predefined labels (e.g., individuals or species) to each audio instance, while detection focuses on identifying subsections of interest in long recordings, often with associated properties like call types.

After extracting audio representations (ht) from the pretrained HuBERT model, we employ mean-pooling to derive a single summary vector for each instance. This vector is then passed through a fully-connected linear layer (fcls) to extract logit values for the classes. For classification,[3] the logits are fed into a softmax layer, and the network is fine-tuned using cross-entropy loss. In detection, we employ a multi-label classification approach with a sigmoid layer, training the network using binary cross-entropy loss. During fine-tuning, we optimize the entire network via gradient-based updates, except for the CNN encoder, which remains frozen, aligning with the approach utilized in HuBERT.

By following this methodology, our research aims to utilize the pretrained representations from HuBERT and apply them effectively to the tasks of classification and detection in the domain of bioacoustics. This approach enables us to employ a unified machine learning model with slight adjustments to the classification layer for different tasks, facilitating knowledge transfer and performance enhancements.

## 4.3    Humane Reinforcement Learning (HRL)

Generating a reward model (RM) calibrated with human preferences is applied. The underlying goal is to get a model or system that takes in a sequence of voice, and returns a scalar reward which should numerically represent the human preference. The system can be an end-to-end LM, or a modular system outputting a reward. The output being a scalar reward is crucial for existing HRL algorithm.

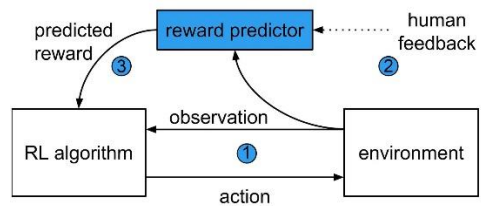

Figure 4.2 Human Feeback based Reinforcement Learning

## 5    Conclusion

To conclude this research offers a significant contribution to the field of animal language construction. The proposed theory of Nonlinear Language Construction using Ism, Fil, and Harf sheds light on the intricate dynamics of animal vocalizations, paving the way for further advancements in understanding and deciphering the rich and diverse languages of our fellow inhabitants on this planet.

> *"Next Time You hear them speak, search for*
> ***Ism*** *and* ***Fil*** *"*

## 7 Foe's Review

*"… Your Theory has the potential to revolutionize our understanding of animal communication … "*

- **Google Bard**

*"… A Perfect 10/10 …"*

- GPT3
Research Preview. ChatGPT May 24 Version